\documentclass[lettersize,journal]{IEEEtran}
\usepackage{amssymb}
\usepackage{amsmath}
\usepackage{mathrsfs}
\usepackage{graphicx}
\usepackage[tight,footnotesize]{subfigure}
\usepackage{subeqnarray}
\usepackage{cite}
\usepackage{color}
\usepackage{extarrows}
\usepackage{mathtools}
\usepackage{dsfont}
\usepackage{float}
\usepackage{enumerate}
\usepackage{marvosym}
\usepackage{upgreek}
\usepackage{url}

\usepackage{multirow}
\usepackage{multicol}
\usepackage{arydshln}
\usepackage{array}
\usepackage{booktabs}  
\usepackage{makecell}

\usepackage[end]{algpseudocode}  
\usepackage{algorithmicx}
\usepackage{algorithm}
\usepackage{url}

\begin{document}
\title{SSN: Stockwell Scattering Network for SAR Image Change Detection}
\author
{Gong Chen, Yanan~Zhao, Yi Wang and Kim-Hui Yap
\thanks
{Manuscript received \today. (Gong Chen and Yanan Zhao contributed equally to this work)  \emph{(Corresponding author: Yi Wang.)}
}
\thanks{Gong Chen is with the Department of Electrical and Computer Engineering, National University of Singapore, Singapore 119077 (e-mail: chengong@u.nus.edu).}
\thanks{Yanan Zhao, Yi Wang and Kim-Hui Yap are with the School of Electrical and Electronic Engineering, Nanyang Technological University, Singapore 639798 (e-mail: YANAN002@e.ntu.edu.sg; wang1241@e.ntu.edu.sg; ekhyap@ntu.edu.sg)}
}
\markboth{IEEE GEOSCIENCE AND REMOTE SENSING LETTERS}%
{Shell \MakeLowercase{\textit{et al.}}: Bare1 Demo of IEEEtran.cls for Journals}

\maketitle
\begin{abstract}
Recently, synthetic aperture radar (SAR) image change detection has become an interesting yet challenging direction due to the presence of speckle noise. Although both traditional and modern learning-driven methods attempted to overcome this challenge, deep convolutional neural networks (DCNNs)-based methods are still hindered by the lack of interpretability and the requirement of large computation power. To overcome this drawback, wavelet scattering network (WSN) and Fourier scattering network (FSN) are proposed. Combining respective merits of WSN and FSN, we propose Stockwell scattering network (SSN) based on Stockwell transform which is widely applied against noisy signals and shows advantageous characteristics in speckle reduction. The proposed SSN provides noise-resilient feature representation and obtains state-of-art performance in SAR image change detection as well as high computational efficiency. Experimental results on three real SAR image datasets demonstrate the effectiveness of the proposed method.
\end{abstract}
\section{Introduction}
\IEEEPARstart{T}{he} past decade has witnessed a widespread growth of interest in image change detection \cite{Radke05} due to its innumerable applications in diverse disciplines including video surveillance \cite{Shahbaz21}, medical diagnosis \cite{Shamul2020}, and remote sensing \cite{Lv2019,Shi2020}, in which change detection in synthetic aperture radar (SAR) has attracted increasing attention in remote sensing communities \cite{Tirandaz2016,Akbarizadeh2012}. However, change detection using SAR images is still a challenging task due to the existence of speckle noise \cite{Davari21}. Therefore, it is essential to develop a robust change detection technique against the speckle noise.

Some pioneering efforts have been made to tackle this issue. Traditionally, the popular methods usually generate a difference image (DI) by comparing multi-temporal images and further analyze the DI to obtain the change map \cite{Celik09,Gong12}. Although some pixel-wise changed information can be captured, it is hard for these methods to exploit the rich feature representations from the original data adaptively. Recently, with the remarkable revolution of deep neural networks, unprecedented performance gains of sensing image change detection methods have been obtained. Wang \emph{et al.} \cite{Wang19} presented a general end-to-end convolutional neural network (CNN) framework for image change detection. Li \emph{et al.} \cite{Li19} proposed a well-designed CNN to learn the spatial characteristics from original images. Gao \emph{et al.} \cite{Gao19} investigated the application of CWNNs in the sea ice change detection, in which the wavelet constrained pooling layer plays an essential role to suppress the speckle noise.
\par The above mentioned deep convolutional neural networks (DCNNs)-based methods have achieved great success by exploiting deep feature representations. However, future development and practical deployment of those methods is hindered by the lack of interpretability and the requirement for massive amounts of training data to deliver the promised performance. To address this issue, the wavelet scattering network (WSN) has been proposed \cite{Mallat12}.
WSN inherits the hierarchical structure of DCNNs, but replace data-driven linear filters with predefined fixed multi-scale wavelet filters. WSN has attracted enormous attention both theoretically and practically. In theory, WSN provides a mathematical understanding of DCNNs \cite{Mallat16} as well as noise-resilient feature representations \cite{Mallat13}. In practice, it offers the state-of-the-art performance in various classification tasks, including handwritten digit recognition \cite{Mallat13}, texture discrimination \cite{Sifre2014}, hyperspectral image classification\cite{Tang15} , especially when the training size is small.
\par However, WSN suffers from a major drawback due to that of wavelet transform: it might not achieve a remarkable performance as expectation because of the rigid multi-scale or algebraic structure as well as the difficulty of choosing appropriate wavelet functions. To solve this problem, Fourier scattering network was proposed by replacing the wavelet basis functions with Gabor ones \cite{Czaja19} which is consistent with simple cells in the mammalian visual cortex biologically and learned filters in neural networks \cite{Czaja20}, and Fourier scattering network (FSN) outperforms WSN in some applications, such as hyperspectral image classification \cite{Tlya20}. Although FSN overcomes the drawback of WSN by using Gabor functions instead of wavelets, it also loses the property of multi-resolution. A natural idea is: why not combine respective merits of FSN and WSN together? A simple solution is to use Stockwell transform to construct the Stockwell scattering network (SSN), since Stockwell transform essentially bridges the gap between short-time Fourier transform and wavelet transform with the multi-resolution in the frequency domain, and it is useful for SAR image despeckling \cite{Fei16}. Besides, SSN inherits the architecture of scattering network which has the property of noise robustness \cite{Mallat12,Mallat13}. Therefore, the objective of this paper is to develop Stockwell scattering network based on Stockwell transform for SAR image change detection. The main contributions of the paper are listed as follows:
 \begin{enumerate}[1)]
 \item We propose the Stockwell scattering network based on three-parameter Stockwell transform, which combines respective advantages of WSN and FSN. SSN provides noise-resilient feature representation and has more flexibility compared with that of WSN and FSN, so that it suppresses speckle noise more efficiently.
  \item To the best of our knowledge, we are the first to introduce SSN into SAR change detection problem. Theoretically, scattering network only requires a small amount of real data obtained in engineering senarios for training. By using the pixel-wise combination framework instead of patches centered at the pixel for training, there is only two images needed for feature extraction in SSN, which is meaningful for decreasing the running costs and computation burden.
  \item Experimental results on three real SAR datasets validate the effectiveness of our proposed Stockwell scattering network for SAR image change detection and promising results are observed.
\end{enumerate}
\begin{figure*}[!t]
\centering
\includegraphics[width=6in]{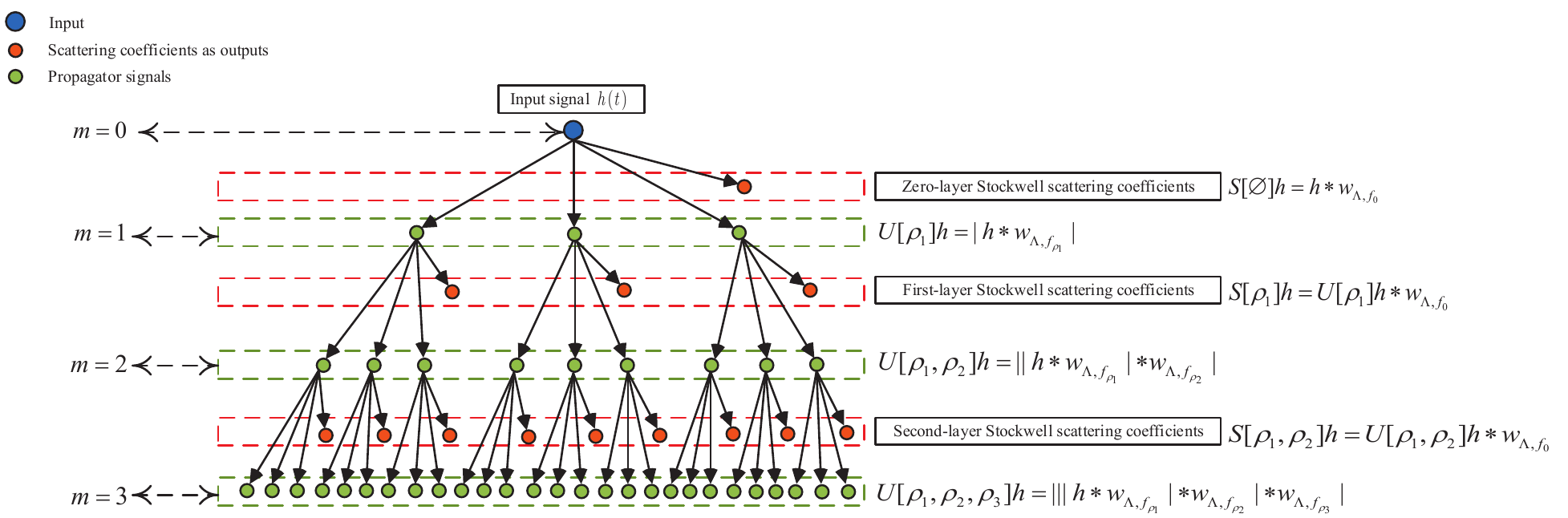}
\caption{Structure of the SSN with $m=3$.}
\label{WSN}
\vspace{-5mm}
\end{figure*}
\section{Proposed Method of stockwell scattering network for SAR images change detection}
In this section, we will introduce Stockwell transform at first, and then we will illustrate how to construct Stockwell scattering network by means of Stockwell transform. Finally, we will further discuss the framework of Stockwell scattering network for SAR images change detection.
\subsection{Overview of Stockwell transform}
The stockwell transform is proposed by Stockwell \emph{et al.} \cite{Stockwell96} to overcome disadvantages of wavelet transform in the choice of wavelets and rigid multi-scale resolution by a hybrid of wavelet transform (WT) and short-time Fourier transform (STFT) with fixed resolution. For the signal $h(t)\in L^{2}(\mathbb{R}^{d})$, the Stockwell transform (ST) is defined as
\begin{equation}
\begin{split}
\label{ST}
S(t,f)=\!\!~<h(\tau),\omega_{t,f}(\tau)>~\!\!=\int_{\mathbb{R}}h(\tau)\!~\omega_{t,f}^{\ast}(\tau)d\tau
\end{split}
\end{equation}
where the transform kernel is expressed as
\begin{equation}
\omega_{t,f}(\tau)=\omega(t-\tau,f)e^{j2\pi f(t-\tau)}
\end{equation}
and $\omega(t-\tau,f)$ is a window function and is generally chosen to be a positive and Gaussian function:
\begin{equation}
\label{window}
\omega(t-\tau,f)=\frac{|f|}{\sqrt{2\pi}}e^{-\frac{\left((t-\tau)f\right)^{2}}{2}}.
\end{equation}
Using the definition of convolution, \eqref{ST} can be also expressed as
\begin{equation}
S(t,f)= h(t)\ast \omega_{t,f}(t).
\end{equation}
Moreover, to improve the time-frequency resolution of ST, the window function can be modified as the three parameters window function in three parameters ST (TPST) \cite{Liu18}:
\begin{equation}
\begin{split}
\label{modify_window}
\omega_{\Lambda}(t-\tau,f)&=\omega_{k,b,c}(t-\tau,f)\\
&=\frac{|kf^{b}+c|}{\sqrt{2\pi}}e^{-\frac{\left(t-\tau\right)^{2}(kf^{b}+c)^{2}}{2}}
\end{split}
\end{equation}
where $\Lambda =(k,b,c)$ is a parameter set, in which $k$, $b$ and $c$ are adjustable parameters, $k$ determines the mode of the window width, $b$ adjusts the changing rate of the window width and $c$ controls the tradeoff between the ST and STFT. It is noticed that if $k=1/a$, $b=1$, $c=0$ ($a$ represents the value of scale), then the TPST reduces to WT, and if $k=0$ and $c\neq 0$, then TPST reduces to STFT. In order to deal with multi-dimensional signal $h(t)\in L^{2}(\mathbb{R}^{d})$, multi-dimensional Stockwell transform with improved window function $\omega_{\Lambda}(t,f)$ is further defined as
\begin{equation}
\begin{split}
\label{ST2}
S(t,f)&=\int_{\mathbb{R}^{d}}h(\tau)\omega_{\Lambda,f}^{\ast}(\tau)d\tau = h(t)\ast \omega_{\Lambda,f}(t)\\
&=\int_{\mathbb{R}^{d}}h(\tau)\omega_{\Lambda}(t-\tau,f)e^{-j2\pi f^{T}(t-\tau)} d\tau
\end{split}
\end{equation}
where
\begin{equation}
\omega_{\Lambda}(t-\tau,f)= \prod_{i=1}^{d}\left(\frac{|kf_{i}^{b}+c|}{\sqrt{2\pi}}e^{-\frac{(t_{i}-\tau_{i})^{2}(kf_{i}^{b}+c)^{2}}{2}}\right)
\end{equation}
\subsection{Stockwell scattering network}
In this section, we are going to construct Stockwell scattering network (SSN) by iterating Stockwell scattering transform which computes noise-robust representation from Stockwell transform followed by non-linear modulus operator.
\par For a signal $h(t)\in L^{2}(\mathbb{R}^{d})$, Stockwell scattering transform computes ST-modulus coefficients, dubbed Stockwell scattering propagation coefficients, by iterating Stockwell transforms and modulus operators. In high dimensional data analysis, it often needs to distinguish data variations along different orientation by introducing directional kernels into the ST in \eqref{ST2}. Denote $\mathbb{G}$ a finite rotation group with elements $r_{n}$, $n\in \{1,2,...,N\}$. It was illustrated in \cite{Mallat12} that if $d$ is even, then $\mathbb{G}$ is a subgroup of the special orthogonal group $\textbf{SO}(d)$; if $d$ is odd, then $\mathbb{G}$ is a subgroup of the orthogonal group $\textbf{O}(d)$. In particular, for 2-D case, $r_{n}$ is a 2-by-2 rotation matrix $r_{n}\triangleq \bigl(\begin{smallmatrix} \cos\theta_{n}&-\sin\theta_{n}\\ \sin\theta_{n}&\cos\theta_{n}\\\end{smallmatrix}\bigr)$ with $\theta_{n}=\frac{2\pi n}{N}$. Directional kernels are derived by rotating the window $\omega_{\Lambda}(t,f)$ along angle $r_{n}\in \mathbb{G}$ and modulating it with $e^{-j2\pi f^{\mathrm{T}}r_{n}t}$:
\begin{equation}
\begin{split}
\label{oriental_window}
\omega_{\Lambda,f,r_{n}}&= \omega_{\Lambda}(r_{n}t,f)e^{-j2\pi f^{\mathrm{T}}r_{n}t}\\
& = \omega_{\Lambda}(t,r_{n}^{-1}f)e^{-j2\pi(r_{n}^{-1}f)^{\mathrm{T}}t}
\end{split}
\end{equation}
in which $\omega_{\Lambda,f_{0}}= \omega_{\Lambda}(t,f_{0})e^{-j2\pi (f_{0})^{\mathrm{T}}t}$, with $f_{0}=0$ corresponding to low frequency, and $\omega_{\Lambda,f_{p},r_{n}}= \omega_{\Lambda}(t,r_{n}^{-1}f_{p})e^{-j2\pi(r_{n}^{-1}f_{p})^{\mathrm{T}}t}$ with high frequencies centered at $f_{p}= p, p\in \mathbb{N}^{+}$. By combining \eqref{oriental_window} and \eqref{ST2}, the coarse approximation and the fine details of $h(t)$ are filtered by $\omega_{\Lambda,f_{0}}$ and by $\omega_{\Lambda,f_{p},r_{n}}$ respectively, then Stockwell coefficients can be computed as following:
\begin{equation}
\begin{split}
S\!~h(t) = \{h\ast \omega_{\Lambda,f_{0}}(t),h\ast \omega_{\Lambda,f_{\rho}}(t)\}_{\rho\in \mathrm{P}}
\end{split}
\end{equation}
where $f_{\rho} = r_{n}^{-1}f_{p}$ and the set $\mathrm{P} =\{\rho=(p,n)|~p\in \mathbb{N}^{+}, n\in \{1,2,...N\}\}$. Then, a Stockwell scattering propagator $\widetilde{S}$ maintains the low-frequency averaging and computes the modulus of the Stockwell coefficients:
\begin{equation}
\widetilde{S}\!~h(t)= \{h\ast \omega_{\Lambda,f_{0}}(t),|h\ast \omega_{\Lambda,f_{\rho}}(t)|\}_{\rho \in \mathrm{P}}.
\end{equation}
For simplicity, the ST-modulus coefficients $|h\ast \omega_{\Lambda,f_{\rho}}(t)|$ are defined as:
\begin{equation}
U[\rho]h=|h\ast \omega_{\Lambda,f_{\rho}}(t)|~~\text{with}~~ \rho \in \mathrm{P},
\end{equation}
and more ST-modulus coefficients can be further computed by iteration of the Stockwell transform and modulus operators. For any ordered sequence $\chi=(\rho_{1},\rho_{2},...,\rho_{m})$, termed as a path with a length of $m$, the Stockwell scattering propagator for a given signal $h(t)$ along the path $\chi$ is defined by cascading ST-modulus operator:
\begin{equation}
\begin{split}
U[\chi]h&=U[\rho_{m}]U[\rho_{m-1}]...U[\rho_{2}]U[\rho_{1}]h\\
&=|~||h\ast\omega_{\Lambda,f_{\rho_{1}}}|\ast\omega_{\Lambda,f_{\rho_{2}}}|\cdots|\ast\omega_{\Lambda,f_{\rho_{m}}}|
\end{split}
\end{equation}
and when $\chi=\emptyset$, $U[\emptyset]h=h$.
In order to perform classification, a Stockwell scattering transform is defined by computing local descriptors with a low-pass filter $\omega_{\Lambda,f_{0}}$:
\begin{equation}
\begin{split}
S[\chi]h&=U[\chi]h\ast\omega_{\Lambda,f_{0}}\\
&=|~||h\ast\omega_{\Lambda,f_{\rho_{1}}}|\ast\omega_{\Lambda,f_{\rho_{2}}}|\cdots|\ast\omega_{\Lambda,f_{\rho_{m}}}|\ast\omega_{\Lambda,f_{0}}
\end{split}
\end{equation}
with $S[\emptyset]h=h\ast\omega_{\Lambda,f_{0}}$. For the path $\chi=(\rho_{1},\rho_{2},...,\rho_{m})$, Stockwell scattering network is constructed by iterating the scattering propagator $\widetilde{W}$. Fig. \ref{WSN} shows the architecture of Stockwell scattering network with three layers. The Stockwell scattering propagator $\widetilde{S}$ is applied to the input signal $h(t)$ to compute the first layer of ST-modulus coefficients $U[\rho_{1}]h$ and output its local average $S[\emptyset]h$. Applying $\widetilde{S}$ to all propagated signals $U[\chi]h$ of the $m$-th layer outputs scattering signals $S[\chi]h$ and computes all propagated signals on the next layer.
\subsection{Classification by Stockwell scattering network for SAR image change detection}
\begin{figure*}[!t]
\centering
\includegraphics[width=5.5in]{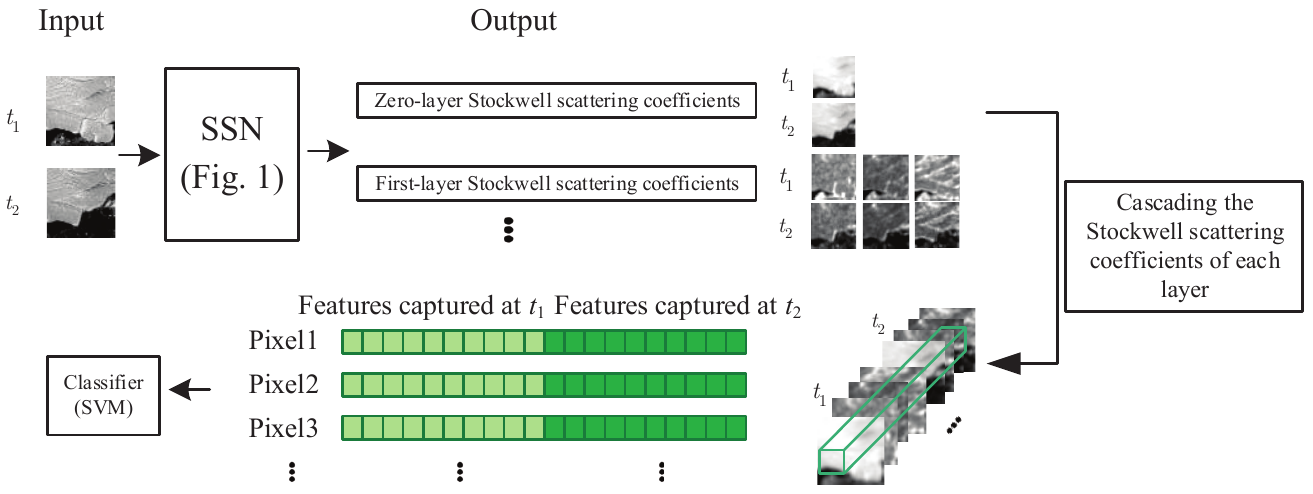}
\vspace{-2.5mm}
\caption{Framework of SSN in SAR images detection.}
\label{Framework}
\vspace{-6mm}
\end{figure*}

Considering two coregistered SAR images $I_{1}$ and $I_{2}$ obtained over the same polar region at different times $t_{1}$ and $t_{2}$, the purpose of SAR images change detection is to produce a difference image that represents the change information between $t_1$ and $t_2$. The framework of SSN applied in SAR image detection is illustrated in Fig. \ref{Framework}.
\par There are mainly four steps. Firstly, Stockwell scattering network with finite number of layers is constructed based on three parameters Stockwell transform. Secondly, two coregistered SAR images $I_{1}$ and $I_{2}$ obtained at different time $t_1$ and $t_2$ are put into the SSN, then Stockwell scattering coefficients (SSCs) from different layers are obtained, which provides noise-robust feature representation of the input SAR images $I_{1}$ and $I_{2}$. Cascading SSCs of $I_{1}$ and $I_{2}$ from each layer, the feature vectors are obtained. For each pixel of SAR images, the corresponding feature vector is composed of the cascade of SSCs from each layer captured at different time $t_{1}$ and $t_{2}$. Finally, classification is then carried out with a Gaussian kernel SVM with the input of feature vectors.
\par In above application process, SSN provides a more efficient computation compared with CWNNs \cite{Gao19}, because there is no need for SSN to create samples from the imaging procedure perspective but an essential step for CWNNs to generate large training samples from original images in order to obtain state-of-art performance.
\section{Experimental Results and Analysis}
\subsection{Experimental Setup}
\begin{figure}[!htbp]
\centering
\includegraphics[width=2.2in]{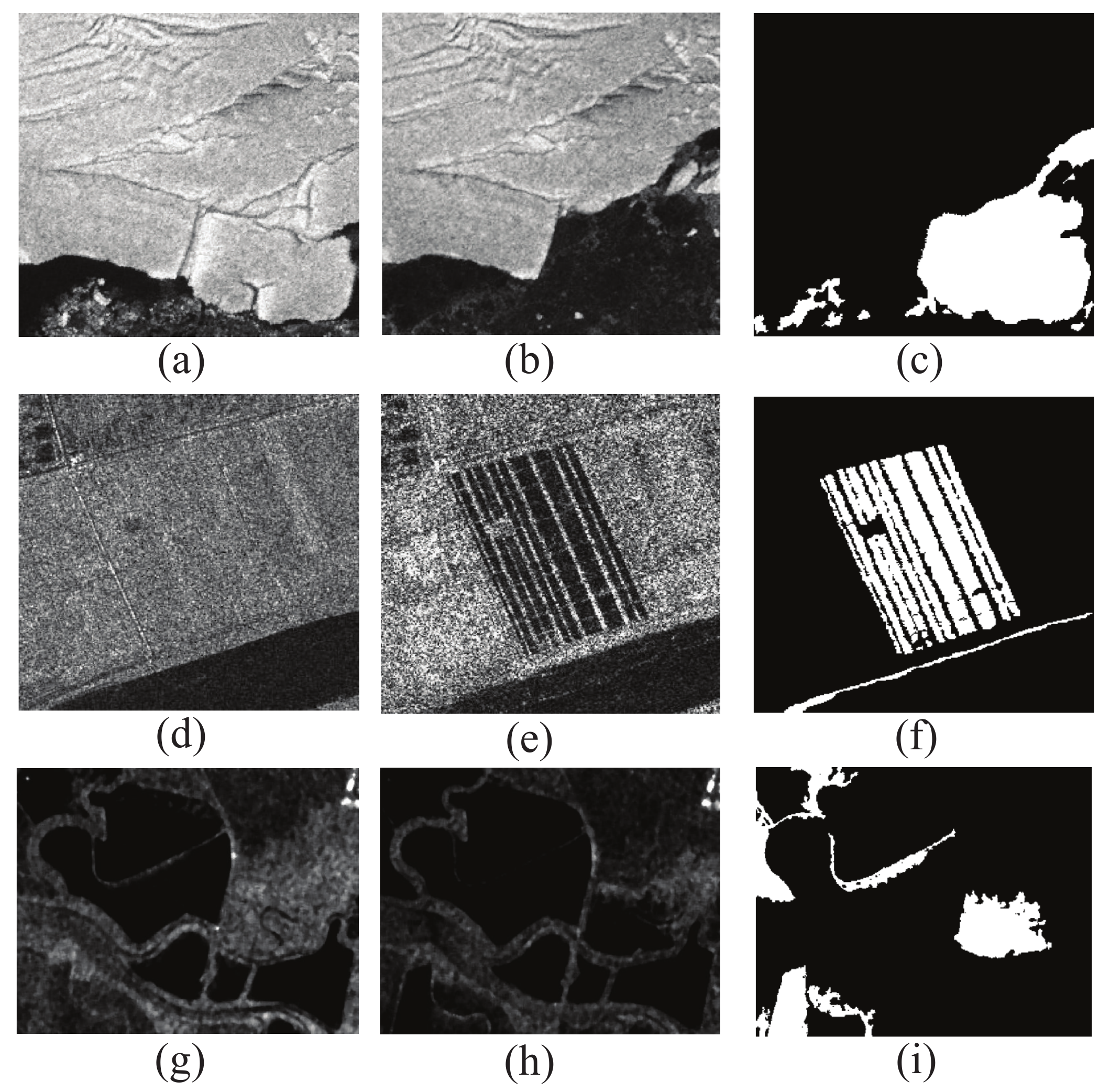}
\vspace{-3.5mm}
\caption{Illustration of the three datasets, first row: Sulzberger dataset, second row: Yellow River dataset, third row: San Francisco dataset. (a) Image acquired in March 11, 2011. (b) Image acquired in March 16, 2011. (c) Ground truth. (d) Image acquired in June 2008. (e) Image acquired in June 2009. (f) Ground truth. (g) Image acquired in August 10, 2003. (h) Image acquired in May 16, 2004. (i) Ground truth.}
\label{Fig_dataset}
\vspace{-8mm}
\end{figure}

To quantitatively evaluate the performance of the proposed method, we use three public datasets acquired by different sensors.
The first one is the Sulzberger dataset with a size of 256$\times$256 pixels, the second one is the Yellow River dataset acquired by the Radarsat-2 satellite with size of 291$\times$306 pixels, and the last one is the San Francisco dataset with a size of 275$\times$400 pixels. Images in last two datasets have strong speckle noise, which are valuable to demonstrate the effectiveness of SSN.
Both images in the datasets and the corresponding ground-truth images are shown in Fig. \ref{Fig_dataset}. Then we evaluate the performance of the proposed SSN by means of six common evaluation metrics \cite{Gao19} in change detection, including false-positives (FP), false-negatives (FN), overall error (OE), percentage of correct classification (PCC), and kappa coefficient (KC) as well as the computation time (CT), in which PCC and KC are essential evaluation metrics, and the larger PCC and KC are, the better performance of the method will obtain.
\par In order to demonstrate the effectiveness of our proposed SSN, we give a comparison of the SSN and several well-known methods (e.g., two traditional methods such as NBRELM \cite{Gao16}, NRCR \cite{Gao18}, three modern learning-based methods including PCANet \cite{PCAnet}, CWNN \cite{Gao19}, and DDNet \cite{Qu22} as well as WSN \cite{Mallat13} and FSN \cite{Czaja20}). For SSN, WSN and FSN, we randomly select 6000 single pixels for training on each dataset (less than 10$\%$), with changed and unchanged samples equal in number. All the methods except for the DDNet are implemented on the same engine with the MATLAB to get the fair comparison of the computation time. DDNet is implemented on the Google Colab platform with supported GPU, so it should compute faster than being put on the MATLAB engine with 8G RAM.
\subsection{Results and analysis}
\begin{figure}[!htbp]
\centering
\includegraphics[width=2.5in]{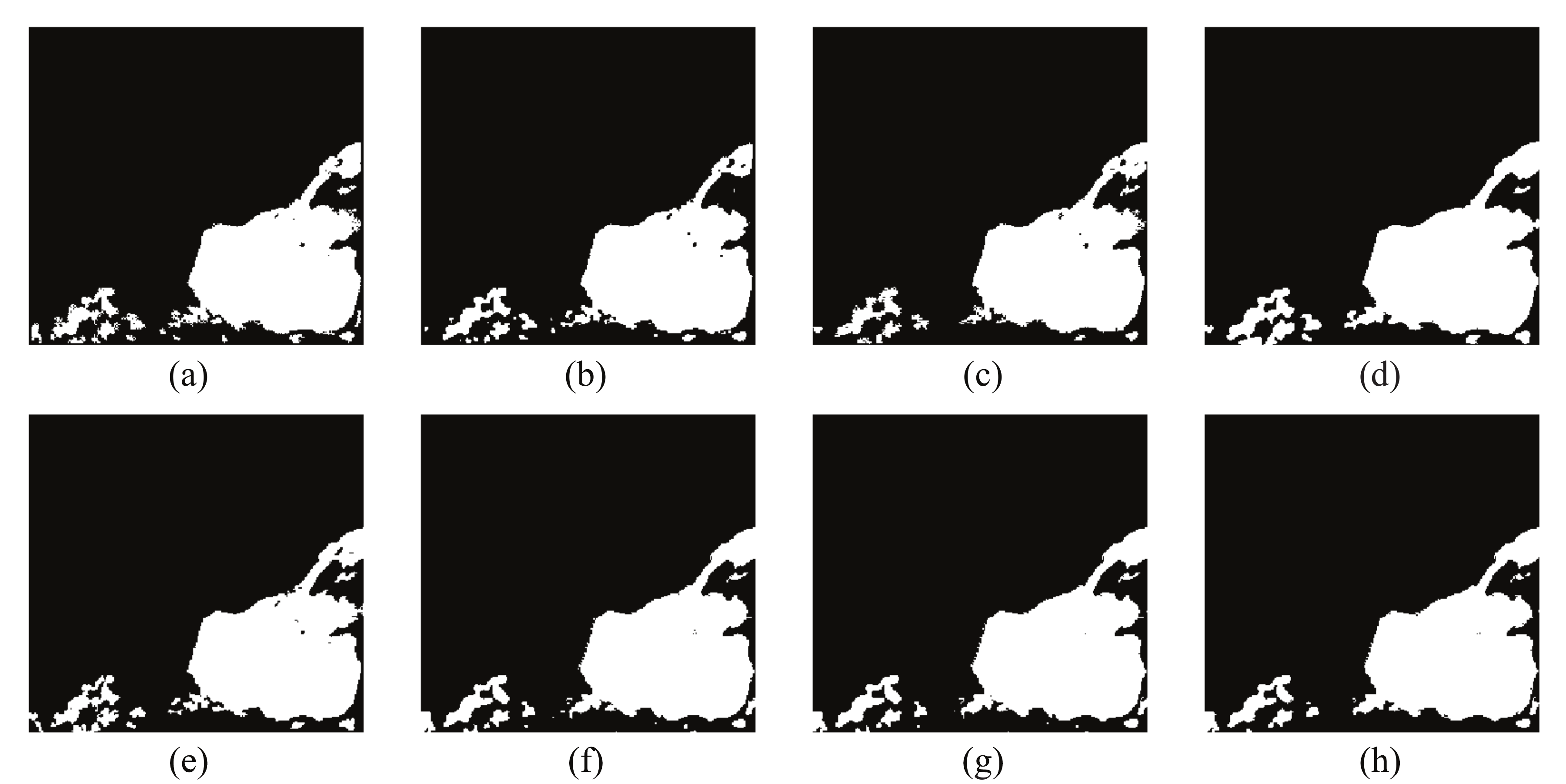}
\vspace{-3.5mm}
\caption{Visualized results of various change detection methods on Sulzberger dataset. (a) Result by NBRELM \cite{Gao16}. (b) Result by NRCR\cite{Gao18}. (c) Result by GaborPCANet \cite{PCAnet}. (d)
Result by CWNN \cite{Gao19}. (e) Result by DDNet\cite{Qu22}. (f) Result by WSN\cite{Mallat13}. (g) Result by FSN\cite{Czaja20}. (h) Result by SSN.}
\label{Fig_sulz}
\vspace{-3mm}
\end{figure}

\begin{figure}[!htbp]
\centering
\includegraphics[width=2.5in]{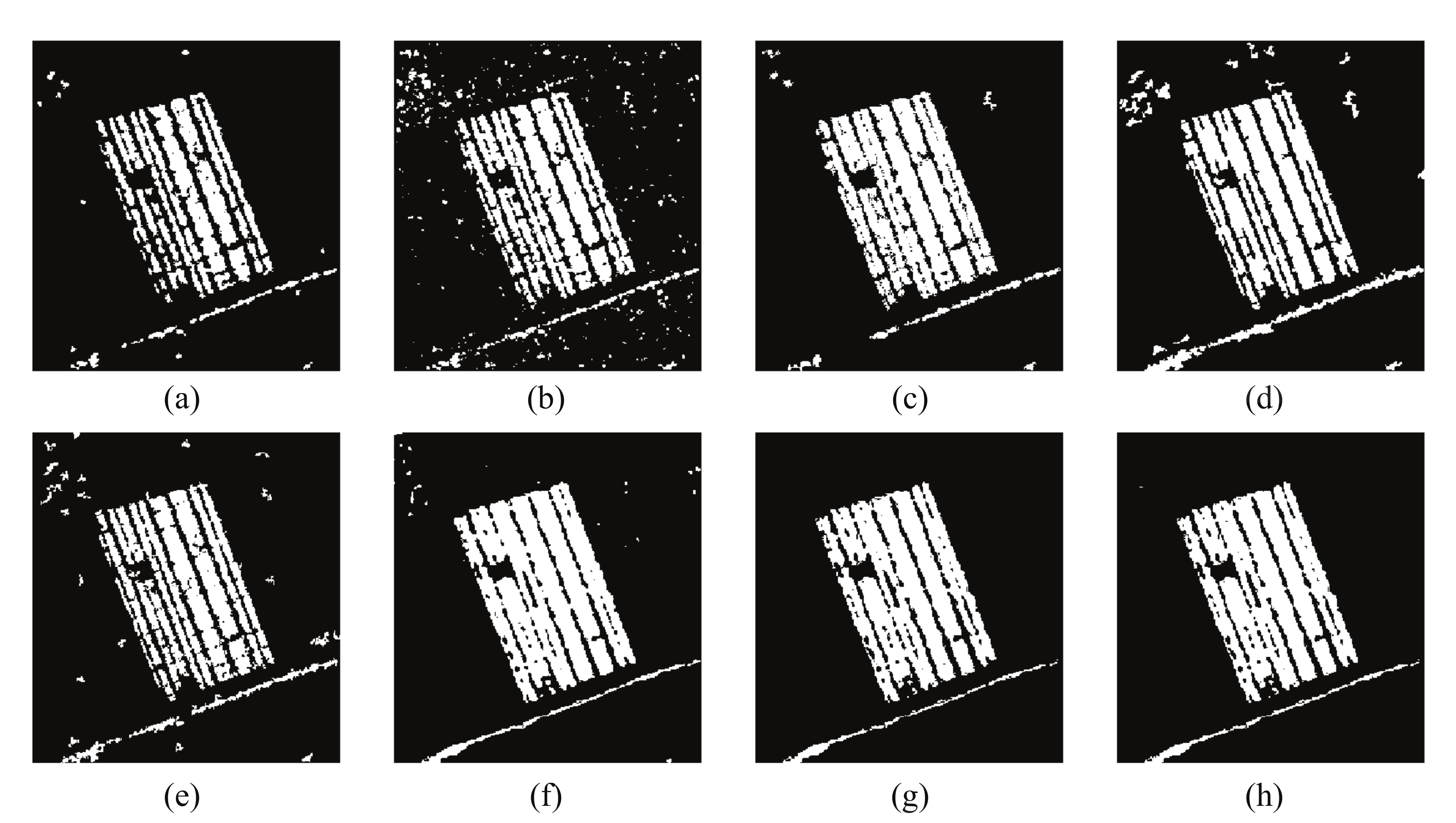}
\vspace{-3.5mm}
\caption{Visualized results of various change detection methods on Yellow River dataset. (a) Result by NBRELM\cite{Gao16}. (b) Result by NRCR \cite{Gao18}. (c) Result by GaborPCANet \cite{PCAnet}. (d) Result by CWNN \cite{Gao19}. (e) Result by DDNet\cite{Qu22}. (f) Result by WSN \cite{Mallat13}. (g) Result by FSN\cite{Czaja20}. (h) Result by the proposed method.}
\label{Fig_yellow2}
\vspace{-3mm}
\end{figure}

\begin{figure}[!htbp]
\centering
\includegraphics[width=2.6in]{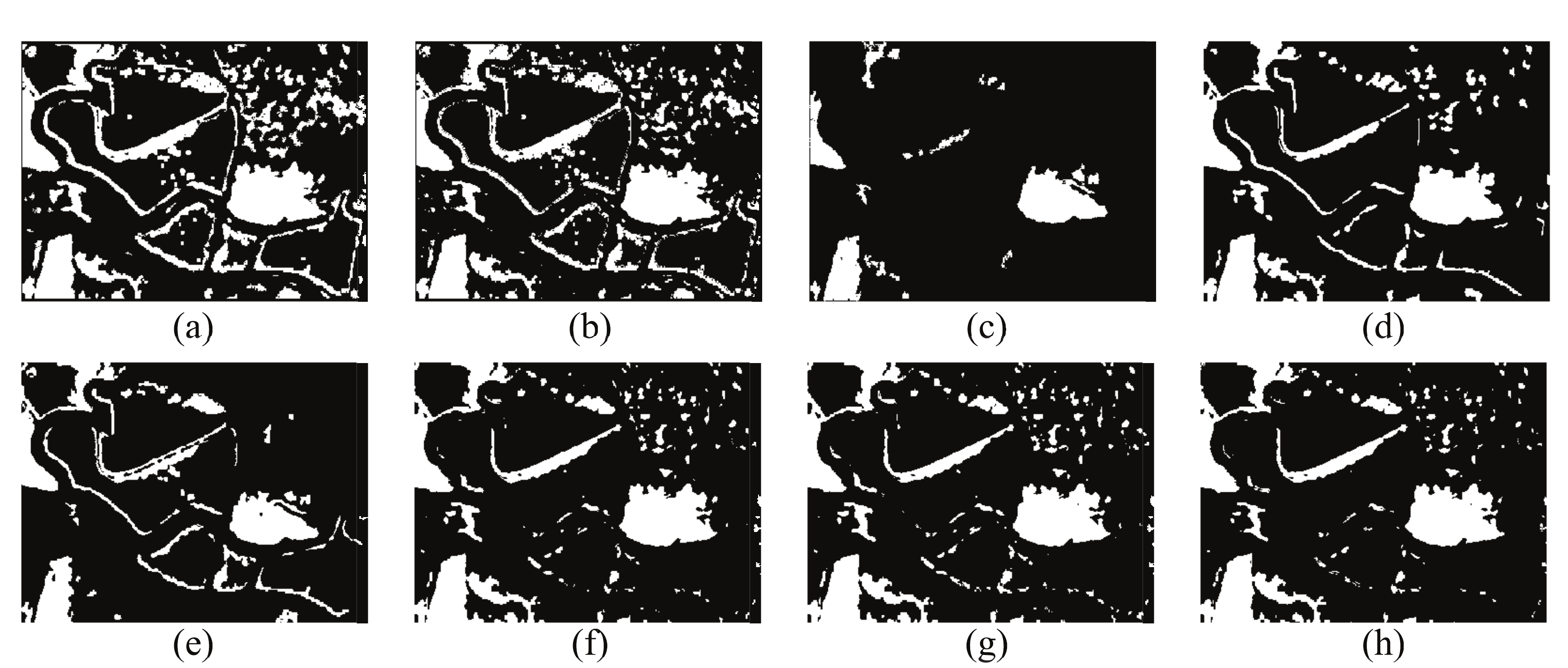}
\vspace{-3.5mm}
\caption{Visualized results of various change detection methods on San Francisco dataset. (a) Result by NBRELM\cite{Gao16}. (b) Result by NRCR \cite{Gao18}. (c) Result by GaborPCANet \cite{PCAnet}. (d) Result by CWNN \cite{Gao19}. (e) Result by DDNet\cite{Qu22}. (f) Result by WSN \cite{Mallat13}. (g) Result by FSN\cite{Czaja20}. (h) Result by the proposed method.}
\label{SF_comparison}
\vspace{-4mm}
\end{figure}

The visualized and quantitative evaluation results of change detection for Sulzberger, Yellow River and San Francisco datasets are shown in Fig. \ref{Fig_sulz}, Fig. \ref{Fig_yellow2}, Fig. \ref{SF_comparison}, Table \ref{Table_sulz}, Table \ref{Table_yellow2} and Table \ref{Table_SF} respectively. Fig. \ref{Fig_sulz}, Fig. \ref{Fig_yellow2} and Fig. \ref{SF_comparison} show the change maps of Sulzberger, Yellow River and San Francisco datasets from different method respectively, in which it is easy to observe that change detected by SSN are much closer to the ground truth illustrated in Fig. \ref{Fig_dataset}.

In order to compare the performance of different methods precisely, Table \ref{Table_sulz} lists the quantitative evaluation results on Sulzberger dataset, from which the proposed SSN outperforms other methods with a PCC of 99.06$\%$. Table \ref{Table_yellow2} and Table \ref{Table_SF} give the quantitative evaluation results on the Yellow River and San Francisco dataset, which shows that PCC and KC values of SSN are obviously better than that of other methods. Besides, the proposed SSN uses the pixel vector with SVM for change detection instead of learning-based patch image training, which largely decreases the computation time. In short, the proposed SSN outperforms other methods both effectively and efficiently.
\begin{table}[!tbhp]
	\renewcommand\tabcolsep{2.8pt}
	\renewcommand\arraystretch{0.8}
	\centering
	\caption{CHANGE DETECTION RESULTS OF DIFFERENT METHODS ON THE Sulzberger DATASET}
    \vspace{-3mm}
	\begin{tabular}{ccccccc}
		\toprule
		Methods & FP &FN&OE& PCC($\%$)&KC($\%$)&CT\\
		\midrule
		NBRELM\cite{Gao16} & 862&356&1218 &98.14 & 94.11 &10.1\\
		\midrule
		NRCR\cite{Gao18}& 508 & 574& 1082 & 98.35& 94.68 &71.7\\
		\midrule
		GaborPCANet\cite{PCAnet}&447&543&990& 98.49& 95.13 &2166.4\\
		\midrule
		CWNN\cite{Gao19}& 1125&228&1353& 97.94& 93.53 &7405.2\\
		\midrule
		DDNet\cite{Qu22}& 754&300&1054& 98.39& 94.90 &286.1$^{*}$\\
		\midrule
		WSN\cite{Mallat13} & 630&82&712& 98.91& 96.56 &38.9\\
		\midrule
		FSN \cite{Czaja20} & 588 & 88 & 676& 98.97& 96.73 &42.3\\
		\midrule
		SSN ($\Lambda =(2.62,1,-0.98)$) & 504 & 109 & 613 & 99.06& 97.03 &35.9\\
		\bottomrule
	\end{tabular}
	*means the computation is acquired by the google colab platform with GPU, which is much more powerful than the authors' engine
	\label{Table_sulz}
    \vspace{-2.5mm}
\end{table}
\begin{table}[!tbhp]
	\renewcommand\tabcolsep{2.8pt}
	\renewcommand\arraystretch{0.8}
	\centering
	\caption{CHANGE DETECTION RESULTS OF DIFFERENT METHODS ON THE Yellow River II DATASET}
    \vspace{-3mm}
	\begin{tabular}{ccccccc}
		\toprule
		Methods & FP &FN&OE& PCC($\%$)&KC($\%$)&CT\\
		\midrule
		NBRELM\cite{Gao16}& 600&3684&4464 & 93.99&77.59 &8.69\\
		\midrule
		NRCR\cite{Gao18}& 2275&2308&4583& 93.83& 79.15&135.6\\
		\midrule
		GaborPCANet\cite{PCAnet}&1885&1581&3466& 95.33& 84.39&3746.5\\
		\midrule
		CWNN\cite{Gao19}& 2076&1198&3274&95.59 & 85.49&3449.0\\
		\midrule
		DDNet\cite{Qu22}& 1290&2215&3505& 95.28&83.63&399.4$^{*}$\\
		\midrule
		WSN\cite{Mallat13} & 1757 & 547 & 2304 & 96.90 & 89.89 & 68.9 \\
		\midrule
		FSN\cite{Czaja20} & 1187 & 1012 & 2199 & 97.04 & 90.06 & 64.3\\
		\midrule
		SSN ($\Lambda =(2.84,1,-0.89)$) & 1292 & 793 & 2085 & 97.19& 90.66&71.3\\
		\bottomrule
	\end{tabular}
	*means the computation is acquired by the google colab platform with GPU, which is much more powerful than the authors' engine
	\label{Table_yellow2}
    \vspace{-4.5mm}
\end{table}
\begin{table}[!tbhp]
	\textcolor{black}{\caption{CHANGE DETECTION RESULTS OF DIFFERENT METHODS ON THE SAN FRANCISCO DATASET}
	\renewcommand\tabcolsep{2.5pt}
	\renewcommand\arraystretch{0.8}
	\centering
	 \label{Table_SF}
    \vspace{-3.5mm}
	\begin{tabular}{ccccccc}
		\toprule
		Methods & FP &FN&OE& PCC($\%$)&KC($\%$)&CT\\
		\midrule
		NBRELM\cite{Gao16}& 15399&746&16145& 85.32&54.29 &11.3\\
		\midrule
		NRCR\cite{Gao18}& 12329& 933 &13262 & 87.94& 59.70&176.3\\
		\midrule
		GaborPCANet\cite{PCAnet}& 390 & 5271 & 5661 & 94.85& 72.72&4870.5\\
		\midrule
		CWNN\cite{Gao19} & 5996  & 1199 & 7195 &93.46 & 74.27&4483.7\\
		\midrule
		DDNet\cite{Qu22}& 5576& 3545 & 9121 & 91.71 & 64.79 &519.2$^{*}$\\
		\midrule
		WSN\cite{Mallat13}& 5351 & 511 & 5862 & 94.67 & 79.06 & 89.5 \\
		\midrule
		FSN\cite{Czaja20}& 5951 & 585 & 6536 & 94.06 & 76.98 & 83.5\\
		\midrule
		SSN ($\Lambda =(2.84,1,-0.89)$) & 4609 & 396 & 5005 & 95.45& 81.82 &92.7\\
		\bottomrule
	\end{tabular}
	*means the computation is acquired by the google colab platform with GPU, which is much more powerful than the authors' engine}
    \vspace{-7mm}
\end{table}
\section{Conclusion}
In this letter, we propose Stockwell scattering network for SAR change detection. Stockwell scattering network is introduced to extract noise-robust features efficiently. It can directly extract features from multi-temporal images without splitting them into numerous patches for learning, which greatly decreases the computation. The experimental results on two datasets show that the proposed Stockwell scattering network can achieve a better performance over several change detection methods with high effectiveness and low computation burden.

\end{document}